\documentclass{article}

\usepackage{preprints}
\usepackage{amsmath}
\usepackage[utf8]{inputenc} 
\usepackage[T1]{fontenc}    
\usepackage{hyperref}       
\usepackage{url}            
\usepackage{booktabs}       
\usepackage{amsfonts}       
\usepackage{nicefrac}       
\usepackage{microtype}      
\usepackage{lipsum}
\usepackage{graphicx}
\graphicspath{ {./images/} }
\usepackage{caption}
\usepackage{needspace}
\usepackage[section]{placeins} 
\usepackage{array} 
\usepackage{multirow}

\title{Quantum-Inspired Evolutionary Neighborhood Search for Arrival-Departure Track Utilization Adjustment under Short-Term Disturbances}
\author{
Xiaobin Li$^*$ \\
  School of Transportation Engineering \\
  East China Jiaotong University\\
  Nanchang, Jiangxi 330038, China   \\
  \texttt{Yuncifor@outlook.com} \\
\And
  Wuming Lei \\
  School of Transportation Engineering \\
East China Jiaotong University\\
  Nanchang, Jiangxi 330038, China   \\
  \texttt{WMFCDS@outlook.com} \\
\And
 Yanbin Gao\\
  School of Transportation Engineering \\
East China Jiaotong University\\
  Nanchang, Jiangxi 330038, China   \\
  \texttt{specoale296@outlook.com} \\
  \And
 Weiguang Wang\\
  School of Transportation Engineering \\
East China Jiaotong University\\
  Nanchang, Jiangxi 330038, China   \\
  \texttt{Wang0422paper@outlook.com} \\
}
\begin{document}
\maketitle
\begin{abstract}
Short-term disturbances at major passenger railway stations alter train arrival and departure times as well as the release sequence of station resources. Effective recovery therefore requires coordinated adjustment of arrival-departure track allocation, station resource occupation, and train retiming. This study represents the station resources involved in train arrival, track occupancy, and departure operations as zone-level resource-occupation intervals. An arrival-departure track allocation adjustment model is formulated. Resource compatibility is imposed as the feasibility condition, while train delays and resource reassignment costs are jointly considered. A quantum-inspired evolutionary algorithm combined with neighborhood search (QEA-NS) is proposed to solve the model.
Perturbation instances are constructed using GTFS timetable data from Frankfurt Hauptbahnhof, Germany. QEA-NS is compared with CP-SAT under the same candidate resource set and feasibility criteria. Both methods generate solutions satisfying the modeled resource compatibility constraints. QEA-NS yields a total delay of 388 min, compared with 519 min for CP-SAT, representing a reduction of 25.2\%. The mean delay of delayed trains decreases from 4.99 to 3.73 min, although QEA-NS requires a longer solution time. Across 10 random perturbation instances, QEA-NS achieves lower total delay in every case. Its mean total delay and standard deviation are 390.5 min and 35.945 min, respectively, compared with 673.8 min and 105.739 min for CP-SAT. The results indicate that, under the adopted resource representation and constraints, QEA-NS improves the delay performance of recovery plans. Its computational efficiency, however, requires further improvement.The implementation is available at \url{https://github.com/yuncifor/QUANTUM-INSPIRED-EVOLUTIONARY-NEIGHBORHOOD-SEARCH-FOR-ARRIVAL-DEPARTURE-TRACK-UTILIZATION}.
\end{abstract}

\section{Introduction}
Major passenger railway stations concentrate train arrival, dwell, and departure operations, and their organizational efficiency directly affects train service regularity. Short-term disturbances, such as primary delays, temporary track restrictions, and extended processing times, can alter train arrival and departure times and the release sequence of station resources. Under high-density operating conditions, local disturbances may propagate to subsequent trains through shared resources, reducing available operating headways and accumulating delays. Therefore, coordinating arrival-departure track allocation, resource occupation, and train retiming while satisfying station-level resource compatibility is a critical problem in disruption recovery at major passenger stations.

Train arrival, track-dwell, and departure operations are temporally connected and occupy corresponding station resources. A shift in train arrival or departure times changes the associated resource-request times and release sequence. When the planned buffer is insufficient, resource occupations may become incompatible with those of other trains. Changes in track assignments or resource combinations may also alter the available resources and operating sequence of subsequent trains. Disruption recovery must therefore determine the required retiming, coordinate track selection and station-resource use, and control both delay levels and the scope of operational adjustments while maintaining resource compatibility.

Prior research on railway disruption response has developed two interrelated streams: train rescheduling and station-resource allocation. Train rescheduling mainly addresses adjustments to train order, routing, and arrival and departure times under disruptions \textsuperscript{\cite{cacchiani2014recovery,corman2015online,ghaemi2017railway}}, with subsequent extensions to robust timetabling \textsuperscript{\cite{lusby2018robustness}} and railway system resilience \textsuperscript{\cite{besinovic2020resilience}}. Station-resource studies focus on the allocation and compatibility of tracks, platforms, and routes. The former primarily aims to restore train service regularity, whereas the latter coordinates the use of station resources. Their decision objects and modeling levels are therefore different. In terms of solution methods, matheuristics combine heuristic search with mathematical programming or local exact optimization and provide useful frameworks for routing and other highly constrained combinatorial optimization problems\textsuperscript{\cite{archetti2014matheuristics,maniezzo2021matheuristics}}. For short-term disruption recovery at major passenger stations, shifts in train times change resource-occupation intervals and release sequences, while resource reallocation affects the feasible operations of subsequent trains. A unified resource-compatibility framework and an appropriate solution method are therefore required to coordinate track selection, station-resource use, and train retiming while controlling delays across feasible recovery plans.

Building on these studies, this paper makes three main contributions. (1) Track, inbound-throat, and outbound-throat resource-occupation intervals are represented uniformly as zone-level resource-occupation intervals. An adjustment optimization model is formulated to integrate conflict detection, feasibility verification, and cost evaluation. (2) A QEA-NS solution method is proposed by combining a quantum-inspired evolutionary algorithm with neighborhood search. QEA explores combinations of train candidate plans through probabilistic observation and improves solution feasibility through conflict-guided adjustment. Starting from a feasible solution obtained by QEA, neighborhood search further reduces total and maximum delays while maintaining zero resource conflicts. (3) A major-station disruption case study is constructed using GTFS timetable data from Frankfurt Hauptbahnhof, Germany. QEA-NS is compared with standalone CP-SAT under the same candidate resource set and feasibility criteria. Multiple random perturbation instances and sensitivity experiments on the number of source trains are used to assess the applicability of the proposed method under different perturbation conditions.

\section{Related work}
\subsection{Train Operation Adjustment and Disruption Response}
Research on railway disruption response initially focused on adjusting train sequences, routes, and arrival and departure times. Törnquist and Persson developed a train rescheduling model for multi-track networks and used discrete decisions to represent train movements under disruptions \textsuperscript{\cite{tornquist2007ntracked}}. Based on alternative graph theory, D’Ariano et al. proposed a branch-and-bound framework \textsuperscript{\cite{dariano2007branch}} and subsequently incorporated train speed coordination into real-time conflict resolution \textsuperscript{\cite{dariano2007conflict}}. To improve computational efficiency in complex scenarios, Corman et al. applied tabu search to train rerouting decisions \textsuperscript{\cite{corman2010tabu}}, whereas Törnquist Krasemann investigated rapid rescheduling from the perspectives of algorithmic structure and computational procedures \textsuperscript{\cite{tornquistkrasemann2012fast}}. These studies established the basic methodological framework for real-time train rescheduling. The main decisions concern train order, routing, and arrival and departure times.

As the disruption scope expands, research has increasingly addressed line blockages and large-scale service disruptions. Meng and Zhou investigated simultaneous train rerouting and rescheduling in multi-track networks and incorporated route selection and timetable adjustment into a unified model \textsuperscript{\cite{meng2014rerouting}}. Louwerse and Huisman developed timetable adjustment methods for partial or complete line blockages \textsuperscript{\cite{louwerse2014blockades}}. Veelenturf et al. formulated a timetable recovery model for large-scale disruptions \textsuperscript{\cite{veelenturf2016largescale}}. Cadarso et al. integrated train service recovery with rolling-stock and passenger-transport operations to study disruption response in rapid transit networks \textsuperscript{\cite{cadarso2013rapid}}. Larsen et al. analyzed the sensitivity of optimal train schedules to stochastic disturbances in process times, showing that changes in disturbance characteristics and parameters can affect the stability of recovery plans\textsuperscript{\cite{larsen2014susceptibility}}. These studies extended train rescheduling from local delay adjustment to line blockages and large-scale disruptions, providing corresponding recovery methods for different disruption scopes.

Further studies have expanded train adjustment problems in terms of both optimization objectives and operational scope. Binder et al. formulated a multi-objective railway timetable rescheduling model that considers trade-offs among different operational performance measures \textsuperscript{\cite{binder2017multiobjective}}. Zhan et al. investigated real-time rescheduling of high-speed trains under complete blockage conditions \textsuperscript{\cite{zhan2015completeblockage}}. Zhu and Goverde expanded the available adjustment strategies under disruptions by incorporating flexible stopping and short-turning \textsuperscript{\cite{zhu2019flexiblestopping}}. Closely related robustness studies aim to reduce the sensitivity of planned operations to uncertainty during the planning stage. These studies include cyclic timetable improvement \textsuperscript{\cite{kroon2008stochastic}}, integrated planning of railway traffic and network maintenance \textsuperscript{\cite{liden2017integrated}}, comparisons between nominal and robust timetables \textsuperscript{\cite{cacchiani2012nominal}}, and the automated construction of robust timetables \textsuperscript{\cite{sels2016robusttimetable}}. Collectively, these studies have broadened the objective system and operational scope of railway adjustment. Overall, existing methods mainly address train order, routing, and arrival and departure times at the network and timetable levels. Station-level arrival-departure track allocation and related resource combinations are generally not adjusted synchronously within the same decision framework.
\subsection{Station-Resource Allocation and Arrival/Departure Route Modeling}
In contrast to network-level train rescheduling, station-level operation studies mainly focus on the allocation relationships among tracks, platforms, and routes. Zwaneveld et al. formulated an integer programming model for route selection through railway stations. They represented the feasibility of candidate plans through resource occupation and route exclusivity constraints \textsuperscript{\cite{zwaneveld1996routing}}. Carey and Carville investigated train scheduling and platform allocation at busy and complex stations, analyzing the coordination between train timetables and platform selection under multi-directional arrival and departure operations \textsuperscript{\cite{carey2003scheduling}}. Lusby et al. systematically reviewed the principal models and solution methods for railway track allocation and examined the relationships among timetable constraints, track selection, and resource conflicts \textsuperscript{\cite{lusby2011trackallocation}}. Caimi et al. applied model predictive control to discrete-time rescheduling in complex central-station areas, extending station-resource allocation to dynamic adjustment scenarios \textsuperscript{\cite{caimi2012predictive}}. These studies developed station-resource allocation and compatibility-assessment methods at different levels of granularity, including tracks, platforms, and routes.

Station-resource allocation and train operation adjustment are interdependent under disruption conditions. A shift in train arrival or departure times changes the occupation intervals and release sequences of the associated resources. Changes in track, platform, or route selection can, in turn, reconfigure the resource-occupation relationships among trains. Existing studies have addressed these interactions from the perspectives of resource allocation, compatibility checking, and dynamic rescheduling, providing a methodological basis for station-level disruption adjustment. Taken together, these studies indicate that station-level recovery must jointly consider changes in train operation times and resource-allocation relationships. Evaluating an adjustment plan from only one decision dimension is therefore insufficient.

\subsection{Quantum-Inspired Evolutionary and Hybrid Optimization Methods}
The quantum-inspired evolutionary algorithm (QEA) was proposed by Han and Kim. Its basic principle is to represent candidate states using probability amplitudes and evolve the population through observation and rotation operations \textsuperscript{\cite{han2002quantuminspired}}. Subsequent studies extended QEA by improving its termination criteria, update operators, and evolutionary procedures \textsuperscript{\cite{han2004termination}}. Zhang systematically reviewed QEA encoding schemes, evolutionary operators, and application areas, and analyzed the effects of different algorithmic designs on the search process \textsuperscript{\cite{zhang2011survey}}. Unlike conventional population representations that directly retain deterministic candidate solutions, QEA describes the selection tendencies of candidate states probabilistically and updates the subsequent search direction according to high-quality solutions obtained during the iterations.

QEA has been applied to various discrete combinatorial optimization problems. Han and Kim applied related methods to classical problems such as the traveling salesman problem and the knapsack problem, demonstrating the applicability of probability-amplitude encoding to combinatorial selection \textsuperscript{\cite{han2000geneticquantum}}. Wang et al. applied a quantum-inspired genetic algorithm to flow-shop scheduling and addressed job sequencing through a combination of probabilistic encoding and classical evolutionary operators \textsuperscript{\cite{wang2005hybrid}}. These studies show that QEA can explore discrete solution spaces involving routing, combinatorial selection, and job sequencing through probabilistic representations. For disruption adjustment at passenger railway stations, the decision structure additionally involves resource selection, temporal coordination, and compatibility constraints. Problem-specific mechanisms for solution representation, evaluation, and improvement are therefore required.

\section{Method}
\subsection{Problem Description and Model Assumptions}
\subsubsection{Problem Description}
This study investigates the adjustment of train arrival-departure track utilization plans under short-term disturbances at passenger railway stations. Given the planned arrival and departure times, candidate tracks, arrival/departure directions, and throat-zone structure, a short-term disturbance can alter train resource-request times, occupation durations, and the release sequence of shared resources. Consequently, some track and route combinations in the original plan may become infeasible.

\begin{figure}[!htbp]
    \centering
    \includegraphics[width=1\linewidth]{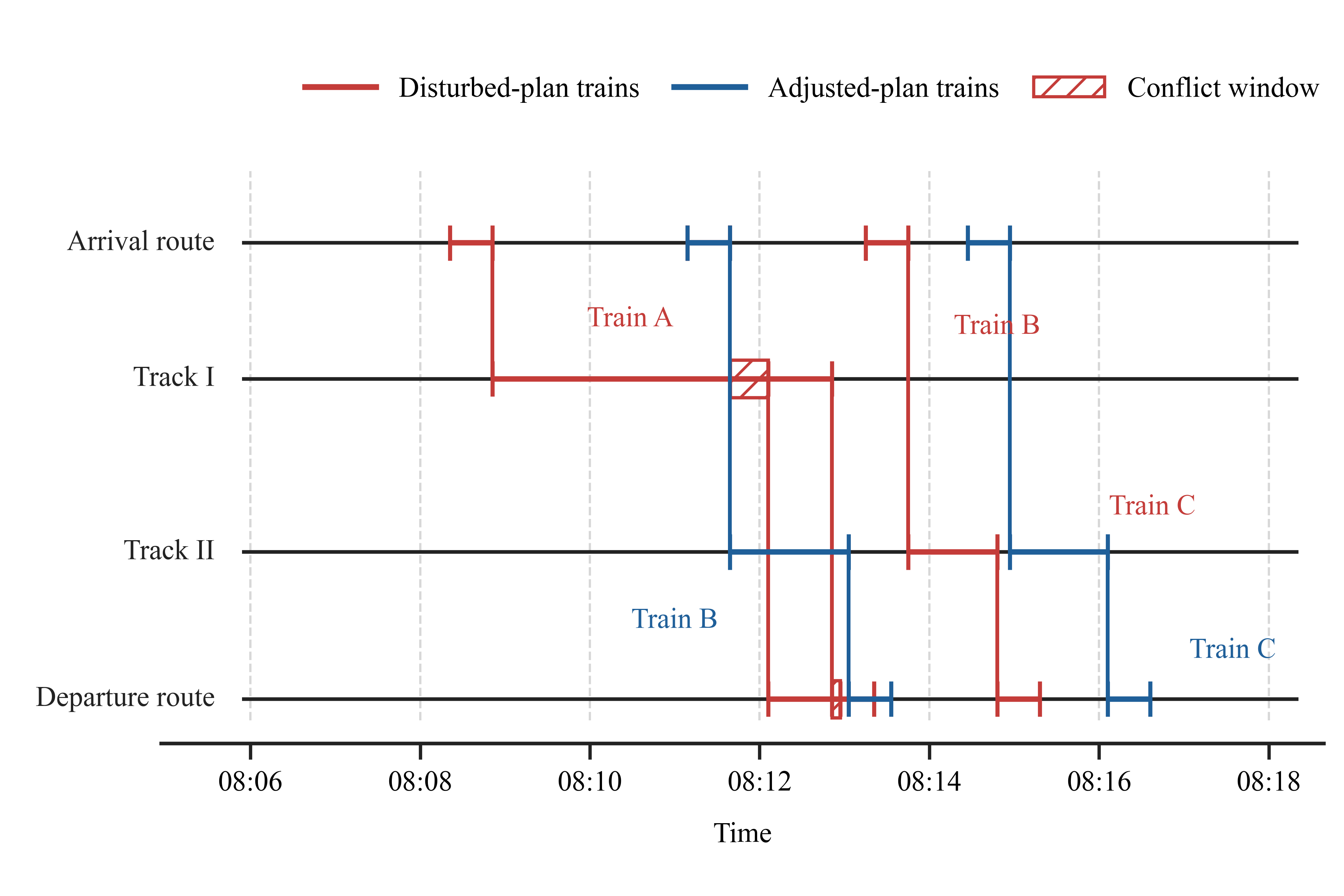}
    \caption{Adjustment of arrival-departure track utilization plan under disturbance}
    \label{fig:Adjustment of arrival-departure track utilization plan under disturbance}
\end{figure}

The adjustment process must redetermine the assigned track, throat passage, and retiming for each train. It must also jointly consider conflict resolution, delay control, and resource reassignment costs to obtain an executable post-disturbance track utilization plan. The adjustment process is illustrated in Fig.\ref{fig:Adjustment of arrival-departure track utilization plan under disturbance}. After a disturbance occurs, the resource-occupation intervals of a scheduled train on its inbound route, arrival-departure track, and outbound route may overlap with those of adjacent trains, resulting in time-window conflicts. During adjustment, conflicting trains can be reassigned to other available arrival-departure tracks, and the operating times of subsequent trains can be moderately retimed. This restores the required safety separations among inbound routes, arrival-departure tracks, and outbound routes.

In this study, a throat passage denotes a resource-based abstraction of the shared passage capacity on the arrival and departure sides of the station for dispatching optimization.

\FloatBarrier

\subsubsection{Model Assumptions}

To facilitate the modeling analysis, the following assumptions are made:
\begin{itemize}
\item The resource-occupation intervals of arrival and departure routes are determined by track location, arrival/departure direction, and the relationships among station operations.
\item Candidate arrival-departure plans are determined by the arrival/departure direction, track adjacency, and reachability of throat passages. Arbitrary route changes that violate station operating rules are not allowed.
\item Throat passages are represented as mutually exclusive resource groups for evaluating the compatibility of resource-occupation intervals during plan adjustment.
\item Train arrival and departure operations, together with the corresponding resource-occupation durations, are specified by model parameters. Short-term disturbances are represented by changes in train resource-request times, temporary unavailability of some resources, or extended occupation durations.
\item The adjustment strategy considers only track and throat-passage reallocation and train retiming. Train cancellations, rerouting, overtaking, changes in arrival/departure direction, and compression of station dwell operations are not considered.
\item Differences in train type, consist length, and traction performance are not distinguished.
\end{itemize}

\subsubsection{Notation}
Table \ref{tab:symbols} summarizes the main sets and variable symbols used in the model formulation.

\begin{table}[htbp]
    \centering
    \caption{Sets and variable notation}
    \label{tab:symbols}
   \begin{tabular}{>{\centering\arraybackslash}m{0.18\linewidth}@{\hspace{2.8em}}>{\centering\arraybackslash}m{0.42\linewidth}}
     \toprule
    \textbf{Symbol} & \textbf{Explanation} \\
    \midrule
    $I$ & Set of trains \\
    $T$ & Set of time instants \\
    $R$ & Set of track and throat passage resources \\
    $P$ & Set of candidate adjustment schemes \\
    $Z$ & Set of resource occupation time intervals \\
    $\Gamma$ & Set of conflicting candidate scheme pairs \\
    \bottomrule
\end{tabular}
\end{table}

\subsection{Short-Term Spatiotemporal Disturbance and Delay Propagation Modeling}
A short-term disturbance scenario consists of four interconnected stages: disturbance-time generation, source-train identification, spatial resource restriction, and propagation under resource constraints. This process maps the original planned operating state to a post-disturbance baseline state that includes initial delays, restricted-resource conditions, and propagation outcomes.

\subsubsection{Disturbance-Time Generation}
Operational density varies across different time periods within a station, leading to heterogeneous disturbance effects. This temporal heterogeneity is characterized by the workloads of train operations, track occupation, and inbound and outbound throat passages. Let \(Q_T(t)\), \(Q_R(t)\), \(Q_S(t)\), and \(Q_H(t)\) denote the normalized workloads of train operations, track occupation, inbound throat passages, and outbound throat passages at time \(t\), respectively. The time-varying disturbance intensity is defined as follows:
\begin{equation}
    \lambda(t) = \omega_T Q_T(t) + \omega_R Q_R(t) + \omega_S Q_S(t) + \omega_H Q_H(t)
\end{equation}
\begin{equation}
\mathrm{Pr}(t_d = t) = \frac{\lambda(t)}{\displaystyle \sum_{\tau\in T}\lambda(\tau)} 
\end{equation}
\begin{equation}
    F(t) = \sum_{\substack{\tau\in T\\\tau\le t}} \mathrm{Pr}(t_d=\tau) 
\end{equation}
\begin{equation}
    t_d= \min\left\{t \,\big|\, F(t)\ge u\right\} 
\end{equation}
\begin{equation}
    T_d = \{t_1,t_2,\dots,t_m\},\quad m \le \bar{n}_T 
\end{equation}
Here, \(\Pr(\cdot)\) denotes the probability of the event in parentheses, and \(F(t)\) denotes the cumulative distribution function up to time \(t\). The variable \(u\) is a random number uniformly drawn from \([0,1]\). The variable \(t_d\) denotes a single disturbance time sampled according to the probability distribution, and \(T_d\) denotes the set of disturbance times selected according to their probability contributions. The parameter \(\bar{n}_T=2\) controls the maximum number of disturbance times. When \(m>1\), disturbance times are sampled without replacement according to \(\Pr(t_d=t)\) until the upper limit \(\bar{n}_T\) is reached. The time-varying workload weights \(\omega_T\), \(\omega_R\), \(\omega_S\), and \(\omega_H\) are set to \(0.35\), \(0.25\), \(0.25\), and \(0.15\), respectively.

\subsubsection{Source-Train Selection and Initial Delay Generation}
\paragraph{Construction of the Source-Train Candidate Set}
After a disturbance time is generated, only trains located within its temporal neighborhood can be directly affected and transmit the impact to subsequent operating chains. For each disturbance time, a source-train candidate set is constructed within its neighborhood as follows:
\begin{equation}
    I_c(t_d) = \{i\in I \mid [s_i,e_i]\cap [t_d-\Delta t,t_d+\Delta t+1]\neq \emptyset\}
\end{equation}
Here $I_c(t_d)$ is the candidate train set corresponding to disturbance time $t_d$; $[s_i,e_i]$ is the station resource occupation time window of train $i$; $\Delta t=3\ \text{min}$ stands for the neighborhood search radius.
\paragraph{Construction of the Source-Train Candidate Set}
Trains in the candidate set may have different impacts on subsequent operations. Source trains are therefore selected from the candidate set according to approximate marginal-impact scores:
\begin{equation}
    \varphi_i \approx \frac{1}{H}\sum_{h=1}^{H}\Delta_i(h)
\end{equation}
\begin{equation}
    I_d(t_d) = \{i\in I_c(t_d) \mid \sigma(i) \le \bar{n_I}\}
\end{equation}
Here, \(\phi_i\) denotes the criticality score of train \(i\); \(H\) denotes the number of random samples; and \(\Delta_i^{(h)}\) denotes the increase in propagation impact caused by inserting train \(i\) in the \(h\)-th random permutation. \(I_d(t_d)\) denotes the set of source trains associated with disturbance time \(t_d\), selected in descending order of \(\phi_i\). The value \(\sigma(i)\) denotes the rank of train \(i\) after sorting all trains in descending order of \(\phi_i\). The parameter \(\bar{n}_I\) controls the number of source trains with initial delays, with \(\bar{n}_I=3\).

After the source trains are identified, their initial delays are generated from a Weibull distribution:
\begin{equation}
v_i\sim\operatorname{Weibull}(\varpi_1,\varpi_2),
\qquad
\delta_i^{0}
=
\min\left\{
\delta_{\max},
\max\left\{1,\operatorname{round}(v_i)\right\}
\right\},
\end{equation}
where \(v_i\) denotes a random sample from the Weibull distribution for train \(i\); \(\varpi_1\) and \(\varpi_2\) are the shape and scale parameters, respectively; and \(\delta_{\max}\) denotes the upper bound of the short-term delay. The parameter values are set to \(\varpi_1=1.65\), \(\varpi_2=2.8\), and \(\delta_{\max}=4\) min.

\subsubsection{Spatial Resource Disturbance Generation}
A reduction in the availability of critical resources provides an effective representation of spatial disturbances. Resource criticality is evaluated for tracks and throat passages based on usage frequency, total occupation duration, peak concurrent occupation, and participation in conflicts. The resources subject to spatial disturbances are then identified according to their criticality scores:
\begin{equation}
R_d(t)=\{r\in R_c(t)\mid \sigma(r)\leq n_R\}
\end{equation}
Here, \(R_c(t)\) denotes the candidate resource set for spatial disturbance, and \(R_d(t)\) denotes the set of resources affected at disturbance time \(t\). The value \(\sigma(r)\) is the rank of resource \(r\) according to its resource-criticality score. The parameter \(n_R=3\) controls the number of disturbed resources. \(B_r\) denotes the restricted duration of resource \(r\), with a range of 2–5 min. \(M_r(\cdot)\) denotes the discrete mapping from the resource-criticality score to the disturbance duration. The score \(\rho_r\) is obtained as a normalized weighted combination of usage frequency, total occupation duration, peak concurrent occupation, and conflict participation. It is used to generate the disturbance intensity of the restricted resource:
\begin{equation}
B_r\sim M_r(\rho_r)
\end{equation}
Track restrictions are represented as temporary unavailability, whereas throat-passage restrictions are represented as increased operating times. Thus, track restrictions affect resource availability, while throat-passage restrictions affect occupation duration. Let \(\mu_r\) denote the slowdown ratio. Then,
\begin{equation}
\mu_r=\mu_{\min}+(\mu_{\max}-\mu_{\min})\rho_r
\end{equation}
\begin{equation}
\tau'_{i,r}=
\begin{cases}
\tau_{i,r}, & r\notin R_d(t),\\
(1+\mu_r)\tau_{i,r}, & r\in R_d(t).
\end{cases}
\end{equation}
Here, \(\mu_{\min}=0.05\) and \(\mu_{\max}=0.15\). The variable \(\tau_{ir}\) denotes the planned occupation duration of resource \(r\) by train \(i\), and \(\tau'_{ir}\) denotes the actual occupation duration under the spatial disturbance.

\subsubsection{Delay Propagation under Resource Constraints}
The initial delays of source trains and restricted resources jointly alter the request and release sequence of station resources. A train successively occupies the corresponding resources along its arrival, dwell, and departure operation chain. Waiting on one resource is propagated to the subsequent resources. Let the resource sequence traversed by train \(i\) be

\begin{equation}
\mathbf{r}_i = \left(r_{i,1}, r_{i,2}, \dots, r_{i,k_i}\right)
\end{equation}
\begin{equation}
\delta_{i,0}=\delta_i^{0}
\end{equation}
denote the initial exogenous delay of train \(i\) before it enters the station-resource chain. During propagation along the above resource sequence, the request time of train \(i\) before entering the \(k\)-th resource is
\begin{equation}
s_{i,k}=s^0_{i,k}+\delta_{i,k-1},
\end{equation}
where \(s^0_{i,k}\) denotes the planned request time for resource \(r_{i,k}\), \(\delta_{i,k-1}\) denotes the cumulative propagation delay before entering the \(k\)-th resource, and \(s_{i,k}\) denotes the request time after incorporating this delay.
When a train requests a resource, the resource may still be occupied by a preceding train. Considering the earliest available time of the resource, the actual entry time is
\begin{equation}
s^*_{i,k}
=
\max\{s_{i,k},t_{r_{i,k}}\},
\end{equation}
where \(t_{r_{i,k}}\) denotes the earliest available time of resource \(r_{i,k}\), and \(s^*_{i,k}\) denotes the actual entry time of train \(i\).
The additional delay caused by resource waiting is
\begin{equation}
w_{i,k}
=
s^*_{i,k}-s_{i,k}
\end{equation}
where \(w_{i,k}\) denotes the additional delay caused by waiting for the resource.
After the train enters the resource, the actual release time is determined by the disturbed occupation duration. The resource availability time is then updated as follows:
\begin{equation}
e^*_{i,k}
=
s^*_{i,k}+\tau'_{i,k},
\qquad
t'_{r_{i,k}}=e^*_{i,k},
\end{equation}
where \(e^*_{i,k}\) denotes the actual release time of the resource, \(t'_{r_{i,k}}\) denotes its updated earliest available time, and \(\tau'_{i,k}\) denotes the occupation duration under the disturbance.
After the resource-release time is updated, the delay caused by waiting continues to propagate to the next resource. The cumulative propagation delay is updated as
\begin{equation}
\delta_{i,k}
=
\delta_{i,k-1}+w_{i,k}
\end{equation}
Here, \(\delta_{i,k}\) denotes the cumulative propagation delay of train \(i\) after passing through the \(k\)-th resource.

\subsection{Resource-Constrained Adjustment Optimization Model for Train Arrival-Departure Track Utilization Plans}
\subsubsection{Candidate Plans and Decision Variables}
The adjustment of train arrival-departure track utilization plans is formulated as a candidate resource-combination selection problem. After a disturbance occurs, the adjustment decision does not change the basic operating chain of any train. Instead, it reselects the track, inbound throat resource, outbound throat resource, and corresponding adjustment plan from a predefined candidate set. The \(n\)-th candidate plan for train \(i\) is defined as
\begin{equation} 
p_{i,n}=
\left(
k_{i,n},
q^{\mathrm{in}}_{i,n},
q^{\mathrm{out}}_{i,n},
z_{i,n}
\right),
\qquad n\in\{1,2,\ldots,m\},
\end{equation}
where \(p_{i,n}\) denotes the \(n\)-th candidate plan for train \(i\), and \(m\) denotes the number of candidate plans. The variable \(k_{i,n}\) denotes the candidate track, while \(q^{\mathrm{in}}_{i,n}\) and \(q^{\mathrm{out}}_{i,n}\) denote the inbound and outbound throat passages, respectively. The set \(z_{i,n}\) contains the resource-occupation intervals associated with the candidate plan.
Each candidate plan specifies a complete set of resource-occupation intervals. Once a train selects a candidate plan, the temporal relationships among its arrival, dwell, and departure operations are determined. For each occupation interval in the candidate plan,
\begin{equation} 
z=
\bigl(
r(z),\tau(z),s(z),e(z)
\bigr),
\qquad z\in z_{i,p},
\end{equation}
where \(r(z)\) denotes the occupied resource, \(\tau(z)\) denotes the operation type, and \(s(z)\) and \(e(z)\) denote the occupation start and end times, respectively.
The binary candidate-plan selection variable is defined as
\begin{equation} 
x_{i,p}=
\begin{cases}
1, & \text{if train } i \text{ selects candidate plan } p,\\
0, & \text{otherwise}.
\end{cases}
\end{equation}
Here, \(x_{i,p}\) indicates whether train \(i\) selects candidate plan \(p\).

\subsubsection{Objective Function}

The adjustment optimization first seeks to eliminate conflicts among resource-occupation intervals. Subject to conflict-free operation, it also aims to control total delay, the maximum adjustment magnitude of individual trains, the number of affected trains, and the scale of resource changes. The objective function is formulated as a weighted composite objective:
\begin{equation}
\min U(x)
=
\eta_1U_1(x)
+\eta_2U_2(x)
+\eta_3U_3(x)
+\eta_4U_4(x)
+\eta_5U_5(x),
\end{equation}
where \(\eta_m\) denotes the weight assigned to the \(m\)-th component objective. The weights are set with sufficiently large separations to establish a priority order. Eliminating resource-occupation interval conflicts has the highest priority, followed by reducing total delay and then controlling maximum delay. The number of adjusted trains and the resource-change cost are minimized jointly at the lowest priority. The component objectives are defined as follows:
\begin{equation}
U_1(x)=\sum_{(i,p,j,p')\in\Gamma}
x_{i,p}x_{j,p'},
\end{equation}
\begin{equation}
U_2(x)=\sum_{i\in I}\delta_i(x),
\end{equation}
\begin{equation}
U_3(x)=\max\{\delta_i(x)\mid i\in I\},
\end{equation}
\begin{equation}
U_4(x) = \sum_{i\in I} \mathbb{I}\left(\delta_i(x)>0\right)
\end{equation}
\begin{equation}
U_5(x)=\sum_{i\in I}\sum_{p\in P_i}g_{i,p}x_{i,p}.
\end{equation}
Here, \(U_1(x)\), \(U_2(x)\), \(U_3(x)\), \(U_4(x)\), and \(U_5(x)\) denote the number of resource-occupation interval conflicts, total delay, maximum delay, number of delayed trains, and resource-change cost, respectively. The function \(\mathbb{I}(\cdot)\) is an indicator function, and \(g_{i,p}\) denotes the resource-change cost of candidate plan \(p\) for train \(i\) relative to the original plan.

\subsubsection{Resource-Occupation Interval Constraints}

Each train must select exactly one candidate track and throat-passage combination in a single adjustment decision:
\begin{equation}
\sum_{p\in P_i}x_{i,p}=1,\qquad i\in I .
\end{equation}
To preserve the temporal continuity of the operating chain, the resource-occupation intervals of the same train during the arrival, dwell, and departure stages are shifted by the same retiming amount. Let \(y_i\in\{0,1,\ldots,20\}\) denote the retiming of train \(i\). The occupation intervals are updated as follows:
\begin{equation}
s_y(z)=s(z)+y_i,\qquad
e_y(z)=e(z)+y_i .
\end{equation}
Here, \(s(z)\) and \(e(z)\) denote the start and end times of occupation interval \(z\) before retiming, while \(s_y(z)\) and \(e_y(z)\) denote the corresponding times after retiming.
To identify occupation conflicts between candidate plans on the same resource, the temporal overlap between two occupation intervals is defined as
\begin{equation}
\chi(z,z')
=
\max\left\{
0,\,
\min\left[e_y(z),e_y(z')\right]
-
\max\left[s_y(z),s_y(z')\right]
\right\}
\end{equation}
Here, \(\chi(z,z')\) denotes the temporal overlap between intervals \(z\) and \(z'\). An overlap constitutes an actual conflict only when the two intervals involve the same resource or mutually exclusive resources. The resource-incompatibility indicator is defined as
\begin{equation}
\kappa(z,z')=
\begin{cases}
1, & r(z)\text{ and }r(z')\text{ are the same or mutually exclusive resources},\\
0, & \text{otherwise}.
\end{cases}
\end{equation}
Here, \(\kappa(z,z')\) indicates whether the resources associated with intervals \(z\) and \(z'\) are incompatible. When temporal overlap and resource incompatibility occur simultaneously, the corresponding candidate plans cannot be selected together. The set of conflicting candidate-plan tuples is defined as
\begin{equation}
\Gamma=
\left\{
(i,p,j,p')
\middle|
i<j,\,
\exists z\in z_{i,p},\,
\exists z'\in z_{j,p'},
\,
\kappa(z,z')=1,\,
\chi(z,z')>0
\right\}.
\end{equation}
Accordingly, the selection constraint for conflicting candidate-plan tuples is
\begin{equation}
x_{i,p}+x_{j,p'}\leq 1,
\qquad
(i,p,j,p')\in\Gamma 
\end{equation}
After the above resource-conflict checks, an adjustment plan is considered feasible if
\begin{equation}
U_1(x)=0 .
\end{equation}
The resource-occupation interval constraints provide a unified check of individual train-plan selection and inter-train competition for shared resources. Feasible plans preserve the continuity of arrival, dwell, and departure operations. Delay and resource-change costs are then compared only among these feasible plans.

\subsection{QEA-NS Solution Algorithm}
\subsubsection{Overall Framework and Parameter Settings}
The QEA-NS algorithm encodes the resource combinations and retiming decisions of the adjustment plans. Its inputs are the post-disturbance resource-occupation state and the set of trains involved in the adjustment. When a candidate plan contains resource-occupation conflicts or excessive retiming, neighborhood search (NS) is invoked to perform local repair around the bottleneck trains.

\begin{figure}[!htbp]
    \centering
    \includegraphics[width=1.2\linewidth]{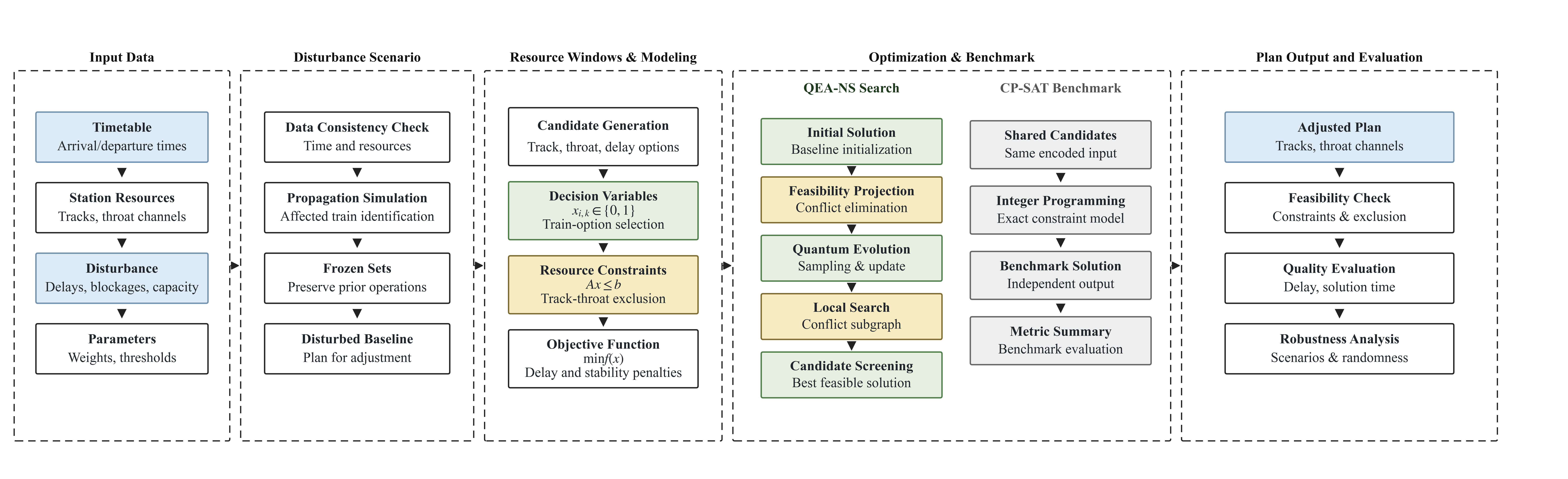}
    \caption{QEA-NS solution framework}
    \label{fig:QEA-NS solution framework}
\end{figure}
\FloatBarrier

\begin{table}[htbp]
  \centering
  \caption{Solution control parameters}
  \label{tab:Solution control parameters}
  \setlength{\tabcolsep}{30pt}
  \begin{tabular}{cc}
    \toprule
    Parameter & Value \\
    \midrule
    Upper limit of basic route candidates for each train & 8 \\
    Delay candidate range & 0--20 min \\
    Upper limit of final candidate schemes for each train & 63 \\
    Population size of QEA & 50 \\
    Maximum iteration number of QEA & 500 \\
    Time limit for QEA stage & 400 s \\
    \bottomrule
  \end{tabular}
\end{table}

\subsubsection{Angle Encoding, Observation, and Updating of Multiple Candidate Plans}

QEA uses candidate plans as the basic encoding units. For train \(i\), each plan \(p\in P\) simultaneously specifies the arrival-departure track, resource combination, and retiming. The angle encoding of an individual in generation \(t\) is
\begin{equation}
\boldsymbol{\theta}^{t}=\left\{\theta_{i,p}^{t}\mid i\in I,\;p\in P\right\},
\qquad
\theta_{i,p}^{t}\in\left[0,\frac{\pi}{2}\right].
\end{equation}
Here, \(\boldsymbol{\theta}^{t}\) denotes the angle encoding of an individual in generation \(t\), and \(\theta_{i,p}^{t}\) denotes the angle parameter associated with candidate plan \(p\) of train \(i\).
Each angle parameter corresponds to a two-dimensional amplitude vector:
\begin{equation}
\left|q_{i,p}^{t}\right\rangle
=
\begin{pmatrix}
\alpha_{i,p}^{t}\\
\beta_{i,p}^{t}
\end{pmatrix}
=
\begin{pmatrix}
\cos\theta_{i,p}^{t}\\
\sin\theta_{i,p}^{t}
\end{pmatrix},
\qquad
\left(\alpha_{i,p}^{t}\right)^2+
\left(\beta_{i,p}^{t}\right)^2=1.
\end{equation}
Here, \(\alpha_{i,p}^{t}\) and \(\beta_{i,p}^{t}\) denote the two amplitude components of candidate plan \(p\).
Because each train must select one plan from multiple candidates, the squared amplitudes of all candidate plans for the same train are normalized:
\begin{equation}
\pi_{i,p}^{t}
=
\frac{\cos^2\theta_{i,p}^{t}}
{\displaystyle\sum_{p'\in P_i}\cos^2\theta_{i,p'}^{t}},
\qquad
\sum_{p\in P_i}\pi_{i,p}^{t}=1.
\end{equation}
Here, \(\pi_{i,p}^{t}\) denotes the observation probability of candidate plan \(p\). A smaller angle produces a larger squared amplitude and, consequently, a higher observation probability. This relationship provides the basis for elite-guided updating.
Given the above probability distribution, the random observation of a candidate plan satisfies
\begin{equation}
\Pr\left(\ell_i^{t}=p\right)=\pi_{i,p}^{t},
\qquad
\end{equation}
where \(\Pr(\cdot)\) denotes the probability of the event in parentheses, and \(\ell_i^{t}\) denotes the index of the candidate plan observed for train \(i\) in generation \(t\). The observation result is then decoded into the model-selection variable:
\begin{equation}
x_{i,p}^{t}
=
\begin{cases}
1, & p=\ell_i^{t},\\
0, & p\neq \ell_i^{t},
\end{cases}
\qquad
\sum_{p\in P_i}x_{i,p}^{t}=1.
\end{equation}
When \(x_{i,p}^{t}=1\), train \(i\) adopts the resource combination and retiming specified by candidate plan \(p\). The selections of all trains form a complete candidate adjustment plan. The corresponding resource-occupation intervals are then reconstructed and the plan is evaluated.
Candidate plans are ranked first by the number of conflicts, followed successively by total delay, maximum delay, number of adjusted trains, and number of resource changes. Better plans are added to the elite candidate set. Let \(\hat{\ell}_i^{t}\) denote the candidate-plan index selected for train \(i\) in the elite-guided plan of generation \(t\). The angle increment is defined as
\begin{equation}
\Delta\theta_{i,p}^{t}
=\begin{cases}
-\gamma_t\theta_{i,p}^{t}, & p=\hat{\ell}_i^{t},\\
\gamma_t\left(\dfrac{\pi}{2}-\theta_{i,p}^{t}\right), & p\neq\hat{\ell}_i^{t}.
\end{cases}
\end{equation}
Here, \(\Delta\theta_{i,p}^{t}\) denotes the elite-guided angle increment for candidate plan \(p\). Its magnitude depends on the distance between the current angle and the target angle. The variable \(\gamma_t\) denotes the update step size in generation \(t\). The candidate selected by the elite plan is moved toward \(0\), whereas the other candidates are moved toward \(\pi/2\).
To balance early exploration and late-stage search stability, the update step size decreases gradually with the number of generations:
\begin{equation}
\gamma_t
=
0.15\left(1-\frac{t}{T^*}\right)+0.05,
\end{equation}
where \(T^*\) denotes the maximum effective number of QEA generations. Thus, the update step size decreases from approximately \(0.20\) to \(0.05\).
The angle parameters in the next generation are determined by the current angles and their increments:
\begin{equation}
\theta_{i,p}^{t+1}
=
\begin{cases}
(1-\gamma_t)\theta_{i,p}^{t}, & p=\hat{\ell}_i^{t},\\
(1-\gamma_t)\theta_{i,p}^{t}+\gamma_t\dfrac{\pi}{2},
& p\neq\hat{\ell}_i^{t}.
\end{cases}
\end{equation}
The updated angles are used for candidate-plan observation in the next generation. Candidate-plan observation, decoding of selection variables, plan evaluation, and elite updating continue until the QEA termination condition is reached, yielding the final selection vector \(x^*\).

\subsubsection{Neighborhood Search and Algorithmic Procedure}

Neighborhood search primarily adjusts trains with positive retiming. The neighborhood train set \(I_b\) is expanded according to the resource-competition relationships associated with candidate plans involving shorter retiming. If the resulting neighborhood is insufficient, trains with similar arrival times are added. During the search, trains outside the neighborhood retain their current plans. Trains in the neighborhood reselect their arrival-departure tracks, resource combinations, and retiming from their respective candidate-plan sets \(P_i\). New plans are compared successively by the number of hard conflicts, total delay, maximum delay, and total number of conflicts. Only an adjustment that outperforms the current plan is accepted.

The QEA-NS procedure is as follows:

\paragraph{Step 1} Generate the disturbance baseline from the original plan and disturbance scenario. Determine the set of trains involved in the adjustment, \(I_a\), and fix the resource-occupation states of trains whose operations were completed before the disturbance.
\paragraph{Step 2} Construct a candidate-plan set \(P_i\) for each train \(i\in I_a\), and generate the initial selection vector \(x^0\) from the disturbance baseline.
\paragraph{Step 3} Initialize the QEA population \(Q^0\). Encode candidate plans using angle encoding and use the current high-quality selection plan to guide the initial distribution of a subset of individuals.
\paragraph{Step 4} In generation \(t\), observe candidate plans and decode the selection variables to obtain a candidate solution \(x^t\). Update the elite solution according to the number of hard conflicts, total delay, maximum delay, number of adjusted trains, and number of resource-combination changes, in that order.
\paragraph{Step 5} Update the angle parameters according to the elite solution and apply candidate-plan adjustments to trains involved in conflicts. Continue until the QEA time budget or the maximum number of iterations is reached.
\paragraph{Step 6} Starting from the feasible solution obtained by QEA, construct the neighborhood train set \(I_b\). Fix the plans of trains outside \(I_b\) and reoptimize the plans of trains within \(I_b\). Compare new solutions according to the number of hard conflicts, total delay, maximum delay, and total number of conflicts, and retain the better solution.
\paragraph{Step 7} After the predefined number of neighborhood-search rounds is completed, select the best feasible solution obtained from QEA and neighborhood search, denoted by \(x^*\).

\section{Case Study Analysis}

\subsection{Case Data and Experimental Design}

The case study is constructed from GTFS arrival and departure records for a weekday in 2024 at Frankfurt Hauptbahnhof, Germany. The station has 29 arrival-departure tracks, with throat passages defined according to the arrival and departure directions. The operation of each train is represented by three resource-occupation intervals: track dwell, inbound throat-passage occupation, and outbound throat-passage occupation.

Four cases are considered for comparison: original-plan verification, the disturbance baseline, standalone CP-SAT, and QEA-NS. CP-SAT and QEA-NS use the same candidate resource set and constraint definitions. At 08:07, trains R005, R001, and R006 incur delays of 2, 1, and 1 min, respectively. Track 26 is blocked for 5 min. Sections W3 and W1 of the western throat experience passage slowdowns lasting 5 and 2 min, respectively, increasing the corresponding arrival and departure operation times of passing trains by 14.7\% and 14.2\%. At 08:08, trains R003, R004, and R007 incur delays of 4, 4, and 2 min, respectively. Tracks 22 and 21 are blocked for 3 and 4 min, respectively, while section E3 of the eastern throat experiences a 4-min passage slowdown. The corresponding arrival and departure operation times of passing trains increase by 12.1\%.

All experiments are conducted on Windows 11 using an Intel Core i9-14900HX processor and 16 GB of memory. The solution environment is Python 3.9.1, and the CP-SAT solver is Google OR-Tools 9.15.

\subsection{Adjustment Feasibility and Conflict Elimination}
A conflict is identified when the resource-occupation intervals of multiple trains overlap on the same resource. This criterion reflects the model-level resource-compatibility check. Under the zone-level throat criterion adopted in this study, a conflict in the original-plan verification case refers to an interval overlap between trains operating within the same throat zone. It does not represent an actual operational route-setting or dispatching conflict. Table \ref{tab:Conflict optimization results of different schemes} summarizes the distribution of resource-occupation interval conflicts across the different plans, while Table \ref{tab:Comparison of adjustment results between QEA-NS and CP-SAT} compares QEA-NS and CP-SAT in terms of feasibility, delay metrics, and solution time.

\begin{table}[htbp]
\centering
\caption{Conflict optimization results of different schemes}
\setlength{\tabcolsep}{12pt}
\begin{tabular}{c c c c c}
\toprule
Index & Original plan review & Disturbance baseline & QEA-NS & CP-SAT \\
\midrule
Inbound throat conflicts & 136 & 124 & 0 & 0 \\
Track occupation conflicts & 0 & 1 & 0 & 0 \\
Outbound throat conflicts & 141 & 131 & 0 & 0 \\
\bottomrule
\end{tabular}
\label{tab:Conflict optimization results of different schemes}
\end{table}
The original-plan verification case in Table \ref{tab:Conflict optimization results of different schemes} contains 277 conflicts. This number reflects the modeling convention used for conflict identification. Publicly available GTFS data do not include switch-level interlocking routes within the station. Therefore, simultaneous occupation within the same throat zone is treated as a conflict and used as an approximate measure of resource competition. The conflicts in the original plan do not indicate a defect in timetable planning. They are a natural consequence of the dense arrangement of station-yard track groups and the shared structure of throat zones.

After the disturbance, one new track-occupation conflict is introduced, whereas the numbers of inbound- and outbound-throat conflicts decrease by 12 and 10, respectively. Consequently, the total number of conflicts in the disturbance baseline decreases from 277 to 256. The disturbance therefore does not increase the total number of conflicts. Instead, it changes the overlap relationships among the original resource-occupation intervals. Some existing throat conflicts are temporally separated, while new track-occupation conflicts arise.

\begin{table}[htbp]
\centering
\caption{Comparison of adjustment results between QEA-NS and CP-SAT}
\setlength{\tabcolsep}{0.5pt} 
\begin{tabular}{c c c c c c c}
\toprule
Method & Computation time / s & Number of conflicts & Total delay / min & Mean delay / min & Maximum delay / min & Number of delayed trains \\
\midrule
QEA-NS & 1293.815 & 0 & 388 & 3.73 & 19 & 104 \\
CP-SAT & 442.717 & 0 & 519 & 4.99 & 22 & 104 \\
\bottomrule
\end{tabular}
\label{tab:Comparison of adjustment results between QEA-NS and CP-SAT}
\end{table}

The total-delay, mean-delay, and maximum-delay metrics in Table \ref{tab:Comparison of adjustment results between QEA-NS and CP-SAT} show that the two methods affect 104 trains each. The numbers of affected trains are therefore comparable, and this dimension does not account for the difference between the two plans. The mean delay is 3.73 min for QEA-NS and 4.99 min for CP-SAT, a difference of 1.26 min. The total-delay gap is not caused by extreme retiming of a small number of trains or by differences in the number of affected trains. Instead, it results from the different levels of control over the retiming of other trains.

The adjustment plans do not simply shift all conflicting trains later. They jointly coordinate train waiting, arrival-departure track selection, and throat-passage use. During periods of limited throat capacity, delay propagates sequentially along the arrival, dwell, and departure operation chain if only retiming is applied. By simultaneously adjusting available tracks and throat passages, the release sequence of resources can be locally reorganized, thereby reducing consecutive waiting times for subsequent trains.

\subsection{Arrival-Departure Track Reallocation and Delay Control}

\begin{figure}[!htbp]
    \centering
    \includegraphics[width=1\linewidth]{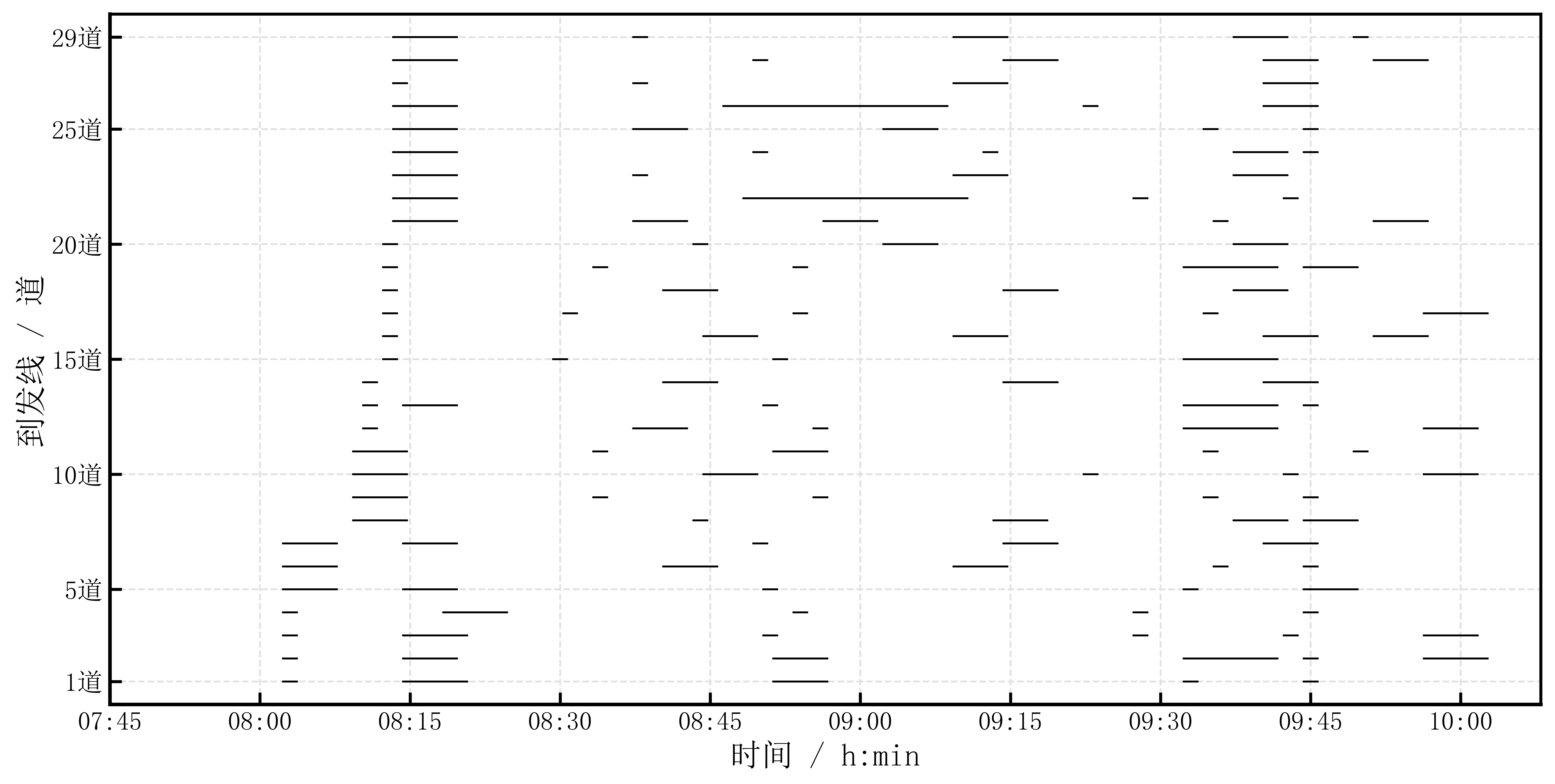}
    \caption{Arrival-departure track occupancy diagram of the original plan}
    \label{fig:Arrival-departure track occupancy diagram of the original plan}
\end{figure}

\begin{figure}[!htbp]
    \centering
    \includegraphics[width=1\linewidth]{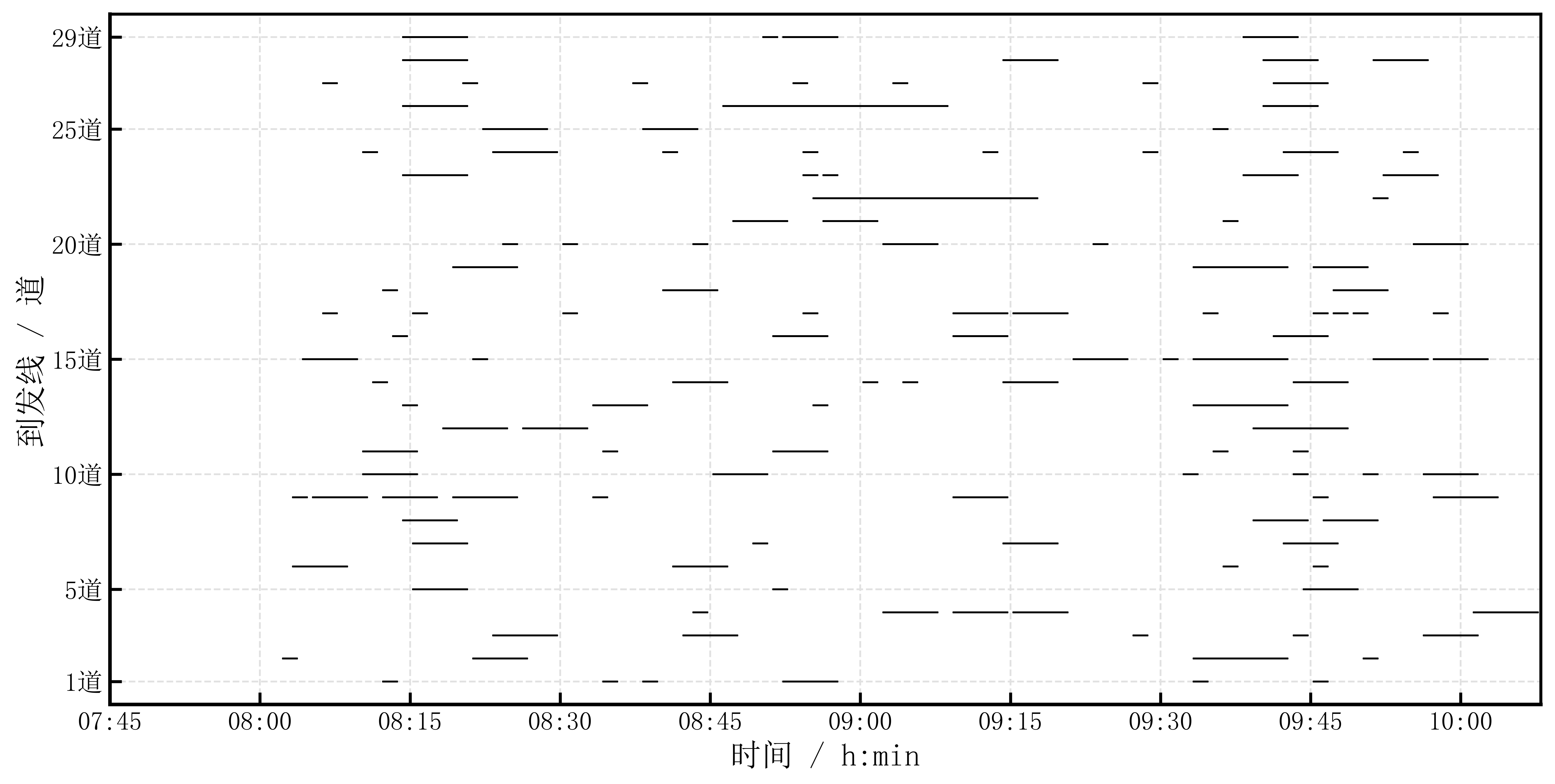}
    \caption{Arrival-departure track occupancy diagram after QEA-NS adjustment}
    \label{fig:Arrival-departure track occupancy diagram after QEA-NS adjustment}
\end{figure}

Arrival-departure track occupation directly reflects the spatiotemporal reconfiguration of train operations after a disturbance. Figure \ref{fig:Arrival-departure track occupancy diagram of the original plan} shows the track-occupation pattern of trains under the original plan during the study period, whereas Fig. \ref{fig:Arrival-departure track occupancy diagram after QEA-NS adjustment} shows the corresponding pattern after QEA-NS adjustment. Table \ref{tab:Adjustment process of representative arrival-departure track occupation} further reports the track assignments and operation time chains of three representative trains under the original, QEA-NS, and CP-SAT plans, illustrating the specific differences in local adjustment decisions.

\begin{table}[htbp]
\centering
\caption{Adjustment process of representative arrival-departure track occupation}
\setlength{\tabcolsep}{9pt}
\begin{tabular}{c c c c c}
\toprule
Scheme & Method & Train & Route & Time chain \\
\midrule
\multirow{3}{*}{Original plan}
& \multirow{3}{*}{--}
& R005 & IN-W-5-E-OUT & 07:51-08:02-08:08-08:14 \\
& & R071 & IN-E-12-W-OUT & 08:46-08:55-08:57-09:04 \\
& & R073 & IN-W-25-E-OUT & 08:55-09:02-09:08-09:19 \\
\cmidrule{1-5}
\multirow{6}{*}{Adjusted scheme}
& \multirow{3}{*}{QEA-NS}
& R005 & IN-W-15-E-OUT & 07:53-08:04-08:10-08:16 \\
& & R071 & IN-E-14-W-OUT & 08:51-09:00-9:02-09:09 \\
& & R073 & IN-W-4-E-OUT & 08:55-09:02-09:08-09:19 \\
\cmidrule{2-5}
& \multirow{3}{*}{CP-SAT}
& R005 & IN-W-9-E-OUT & 07:53-08:04-08:10-08:16 \\
& & R071 & IN-E-12-W-OUT & 08:46-08:55-08:57-09:04 \\
& & R073 & IN-W-17-E-OUT & 08:56-09:03-09:09-09:20 \\
\bottomrule
\end{tabular}
\label{tab:Adjustment process of representative arrival-departure track occupation}
\end{table}

\FloatBarrier

\FloatBarrier

Note: In the routes, W and E denote the west and east throat passages, respectively. The time chain is listed in the sequence of inbound start, arrival, departure and outbound end.

The comparison between Figs. \ref{fig:Arrival-departure track occupancy diagram of the original plan} and \ref{fig:Arrival-departure track occupancy diagram after QEA-NS adjustment} shows that the difference between the two plans is not manifested as a uniform shift of all train times within the study period. Instead, it is concentrated on the reorganization of track assignments. When a disturbance occurs during a high-density operating period, the adjustment does not simply shift all subsequent trains later, which would be the simplest but most coarse-grained strategy. Instead, track assignments are reorganized within the high-density period around the source trains.

The adjustment patterns in Table \ref{tab:Adjustment process of representative arrival-departure track occupation} show that both QEA-NS and CP-SAT reassign multiple trains among throat zones on the same side. For example, the track assigned to R073 is changed from Track 25 in the original plan to Track 4 or Track 17, while the track assigned to R071 either remains Track 12 or changes to Track 14. Different throat zones correspond to physically separated operating areas within the station yard. Distributing concurrent operations originally assigned to the same zone across adjacent zones effectively expands the spatial utilization of instantaneous arrival and departure capacity. This allows conflicts to be absorbed without substantially shifting the timetable.

\subsection{Stability across Disturbance Scenarios and Sensitivity to the Number of Source Trains}

\begin{figure}[!htbp]
    \centering
    \includegraphics[width=1\linewidth]{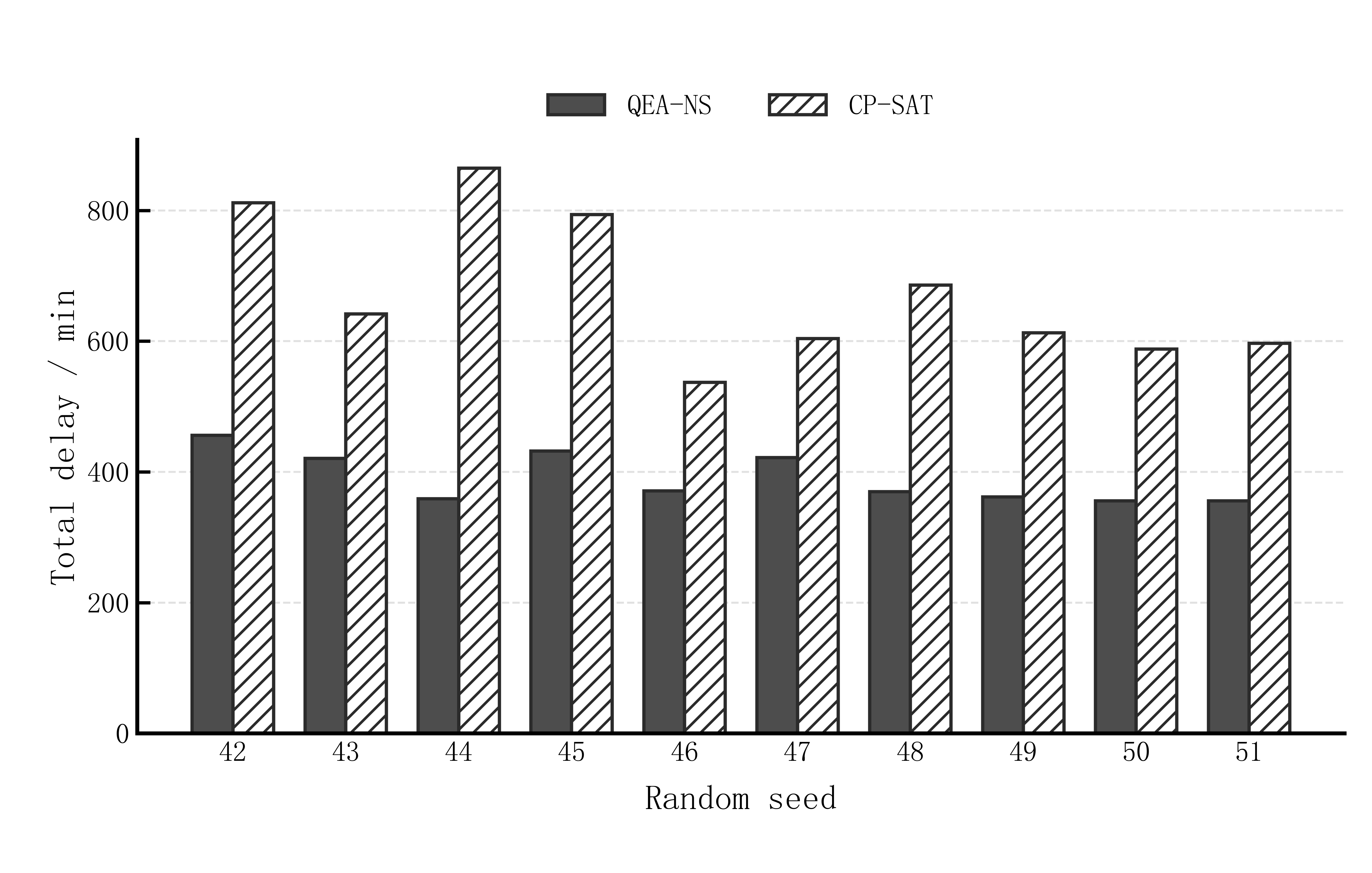}
    \caption{Comparison of total delay under multiple random seeds}
    \label{fig:Comparison of total delay under multiple random seeds}
\end{figure}

To assess the applicability of QEA-NS under different disturbance conditions, only the number of source trains selected according to their scores at each disturbance time is varied. The stability of QEA-NS and CP-SAT solution quality across different random disturbance samples is evaluated, with the results presented in Table \ref{tab:Statistics of adjustment effects under different disturbance samples} and Fig. \ref{fig:Comparison of total delay under multiple random seeds}. 

\begin{table}[htbp]
  \centering
  \caption{Statistics of adjustment effects under different disturbance samples}
  \label{tab:Statistics of adjustment effects under different disturbance samples}
  \setlength{\tabcolsep}{30pt}
  \begin{tabular}{ccc}
    \toprule
    Index & QEA-NS & CP-SAT \\
    \midrule
    Average total delay / min & 390.5 & 673.8 \\
    Median total delay / min & 370.5 & 627.5 \\
    Standard deviation of total delay / min & 35.945 & 105.739 \\
    Average computation time / s & 1170.589 & 211.175 \\
    \bottomrule
  \end{tabular}
\end{table}

\FloatBarrier

\begin{figure}[!htbp]
    \centering
    \includegraphics[width=1\linewidth]{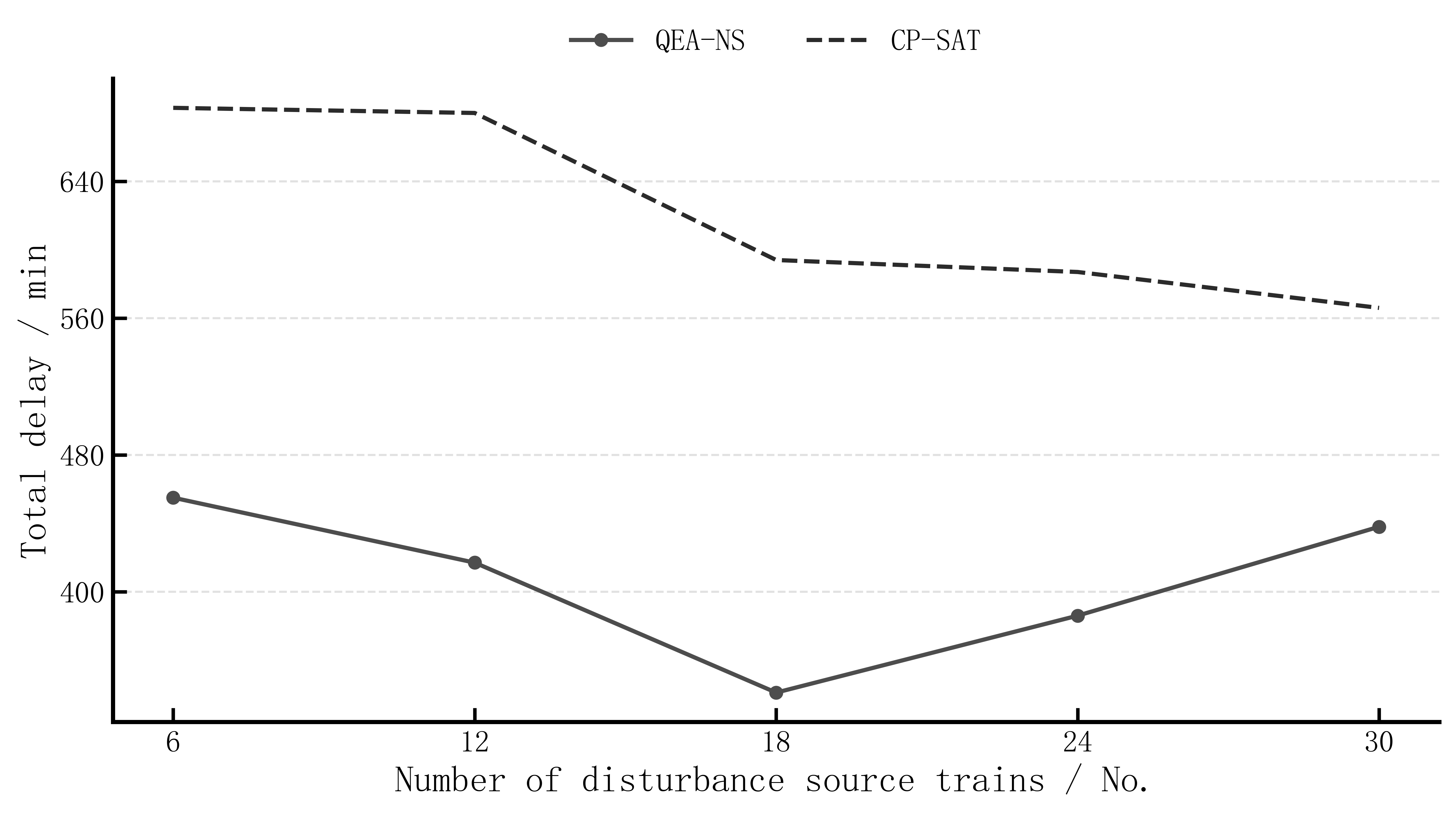}
    \caption{Sensitivity analysis of the number of disturbed source trains}
    \label{fig:Sensitivity analysis of the number of disturbed source trains}
\end{figure}
\FloatBarrier
Table \ref{fig:Comparison of total delay under multiple random seeds} shows that QEA-NS obtains conflict-free solutions for all 10 random disturbance samples. Its mean total delay is 390.5 min, compared with 673.8 min for CP-SAT, and its standard deviation is 35.945 min, compared with 105.739 min for CP-SAT. QEA-NS outperforms CP-SAT for all 10 random seeds. Its relative performance remains consistent across disturbance instances and is not driven by any particular random seed.

Figure \ref{fig:Sensitivity analysis of the number of disturbed source trains} shows that QEA-NS consistently achieves lower total delay than CP-SAT across the five settings, in which the number of source trains increases from 6 to 30. The total delay of QEA-NS decreases from 455 min at 6 source trains to 341 min at 18 source trains, and then increases to 386 and 438 min at 24 and 30 source trains, respectively. CP-SAT reaches 683 min at 6 source trains and then decreases overall to 566 min at 30 source trains. The total-delay gaps between QEA-NS and CP-SAT are 228, 263, 253, 201, and 128 min, respectively. These results indicate that increasing the number of source trains does not change total delay in a fixed proportion. Across the entire tested range, however, QEA-NS consistently obtains lower total delay than CP-SAT.

\section{Conclusions}
This study addresses the adjustment of arrival-departure track utilization plans under short-term disturbances at major passenger railway stations. Track occupation, arrival-departure routes, and train retiming are incorporated into a unified multidimensional optimization framework. Simultaneous occupation within the same throat zone is included in the conflict definition. Compared with formulations based only on track occupation or minimum headway checks, the proposed resource-occupation interval constraints jointly account for arrival/departure directions, throat sides, and shared track-group structures in the compatibility assessment. This prevents plans that appear feasible based on track availability but cannot be implemented because of occupied throat capacity. The proposed modeling framework provides a computable basis for jointly deciding track allocation and retiming under high-density station operations.

Under the same candidate resource set and constraint definitions, both QEA-NS and CP-SAT eliminate all 256 resource-occupation interval conflicts in the disturbance baseline. In the benchmark case, QEA-NS obtains a total delay of 388 min and a mean delay of 3.73 min among delayed trains, compared with 519 min and 4.99 min for CP-SAT. The maximum individual delays are 19 and 22 min, respectively. Both methods reassign multiple trains within the candidate track pool but converge to different adjustment combinations. This result indicates that the feasible region under the given candidate set and constraints contains multiple structurally distinct recovery plans. Track reassignment can help resolve conflicts and produce a more balanced delay distribution.

In the multiple-disturbance experiments, QEA-NS achieves a mean total delay of 390.5 min and a standard deviation of 35.945 min across 10 random disturbance samples. Both values are lower than those of CP-SAT, which are 673.8 and 105.739 min, respectively. This stable performance difference indicates that the advantage of QEA-NS does not result from a particular disturbance sample. In the sensitivity analysis of the number of source trains, the delay gap between the two methods varies non-monotonically with disturbance scale. This finding suggests that disturbance scale is only one factor determining the relative performance of recovery methods. The specific resource-competition pattern generated by each disturbance structure also affects the adjustment outcome.

This study has several limitations. It considers short-term disturbances at a single major passenger station. Throat passages are represented as zone-level aggregated resources rather than switch-level interlocking route topologies. Cross-station train interactions, real-time passenger-flow feedback, and integrated conflicts involving maintenance windows and shunting operations are not considered. Future research should incorporate actual interlocking-route data and historical disturbance records to compare models under switch-level conflict detection. Dynamic resource-state models and adaptive adjustment strategies should also be developed for multiple and uncertain disturbance sources.
\section*{Code Availability}
The source code and README files used in this study are publicly available at:
\url{https://github.com/yuncifor/QUANTUM-INSPIRED-EVOLUTIONARY-NEIGHBORHOOD-SEARCH-FOR-ARRIVAL-DEPARTURE-TRACK-UTILIZATION}
\bibliographystyle{unsrt}
\bibliography{references}  

\end{document}